\documentclass[conference]{IEEEtran}
\IEEEoverridecommandlockouts
\usepackage{cite}
\usepackage{amsmath,amssymb,amsfonts}
\usepackage{graphicx}
\usepackage{textcomp}
\usepackage{xcolor}
\usepackage{url}
\usepackage{caption}
\usepackage{subcaption}
\usepackage{float}
\usepackage{algorithm}
\usepackage[noend]{algpseudocode}
\usepackage{booktabs}

\usepackage[letterpaper,top=0.8in,bottom=1.05in,left=1.05in,right=1.05in]{geometry}
\usepackage{microtype}          
\def\BibTeX{{\rm B\kern-.05em{\sc i\kern-.025em b}\kern-.08em
    T\kern-.1667em\lower.7ex\hbox{E}\kern-.125emX}}
\begin{document}

\title{Synthetic Swarm Mosquito Dataset for Acoustic Classification: A Proof of Concept\\
}

\author{\IEEEauthorblockN{1\textsuperscript{st} Thai Duy Dinh}
\IEEEauthorblockA{\textit{Faculty of Engineering} \\
\textit{Vietnamese-German University}\\
Ho Chi Minh City, Vietnam \\
10421014@student.vgu.edu.vn}
\and
\IEEEauthorblockN{2\textsuperscript{nd} Minh Luan Vo}
\IEEEauthorblockA{\textit{Faculty of Engineering} \\
\textit{Vietnamese-German University}\\
Ho Chi Minh City, Vietnam \\
10423074@student.vgu.edu.vn}
\and
\IEEEauthorblockN{Cuong Tuan Nguyen  \IEEEauthorrefmark{2}}
\IEEEauthorblockA{\textit{Faculty of Engineering} \\
\textit{Vietnamese-German University}\\
Ho Chi Minh City, Vietnam \\
cuong.nt2@vgu.edu.vn}
\and
\IEEEauthorblockN{Hien Bich Vo \IEEEauthorrefmark{2} }
\IEEEauthorblockA{\textit{Faculty of Engineering} \\
\textit{Vietnamese-German University}\\
Ho Chi Minh City, Vietnam \\
hien.vb@vgu.edu.vn}
\and
\thanks{\IEEEauthorrefmark{2} are the correspondence authors.}
}

\maketitle

\begin{abstract}
Mosquito-borne diseases pose a serious global health threat, causing over 700,000 deaths annually. This work introduces a proof-of-concept Synthetic Swarm Mosquito Dataset for Acoustic Classification, created to simulate realistic multi-species and noisy swarm conditions. Unlike conventional datasets that require labor-intensive recording of individual mosquitoes, the synthetic approach enables scalable data generation while reducing human resource demands. Synthetic swarm audio generation is the core novelty of this work. This approach facilitates the development of realistic, scalable multi-species datasets that would be impractical to collect through fieldwork. Using log-mel spectrograms, we evaluated lightweight deep learning architectures for the classification of mosquito species. Experiments show that these models can effectively identify six major mosquito vectors and are suitable for deployment on embedded low-power devices. The study demonstrates the potential of synthetic swarm audio datasets to accelerate acoustic mosquito research and enable scalable real-time surveillance solutions. The public dataset used in this study can be found here \footnote{\url{https://github.com/duydinhthai27/synthetic-swarm-mosquito}}
\end{abstract}


\section{Introduction}
Mosquito-borne diseases—including dengue, malaria, and Zika—infect hundreds of millions each year, with key vectors such as \textit{Aedes aegypti}, \textit{Anopheles}, and \textit{Culex} driving transmission \cite{breedlove2022deadly}. Their growing range, fueled by urbanization and climate change, underscores the need for scalable surveillance.

Conventional methods like trapping and lab identification are slow and labor-intensive. These limits were evident during the 2015–2016 Zika outbreak, where delayed vector detection worsened the spread \cite{boyer2018overview}.

Audio-based detection via wingbeat frequency offers a low-cost, non-invasive alternative, but most systems are built for clean environments and single-species detection—limiting real-world applicability on embedded hardware.

\begin{figure}[htbp]
    \centering
    \includegraphics[width=\linewidth]{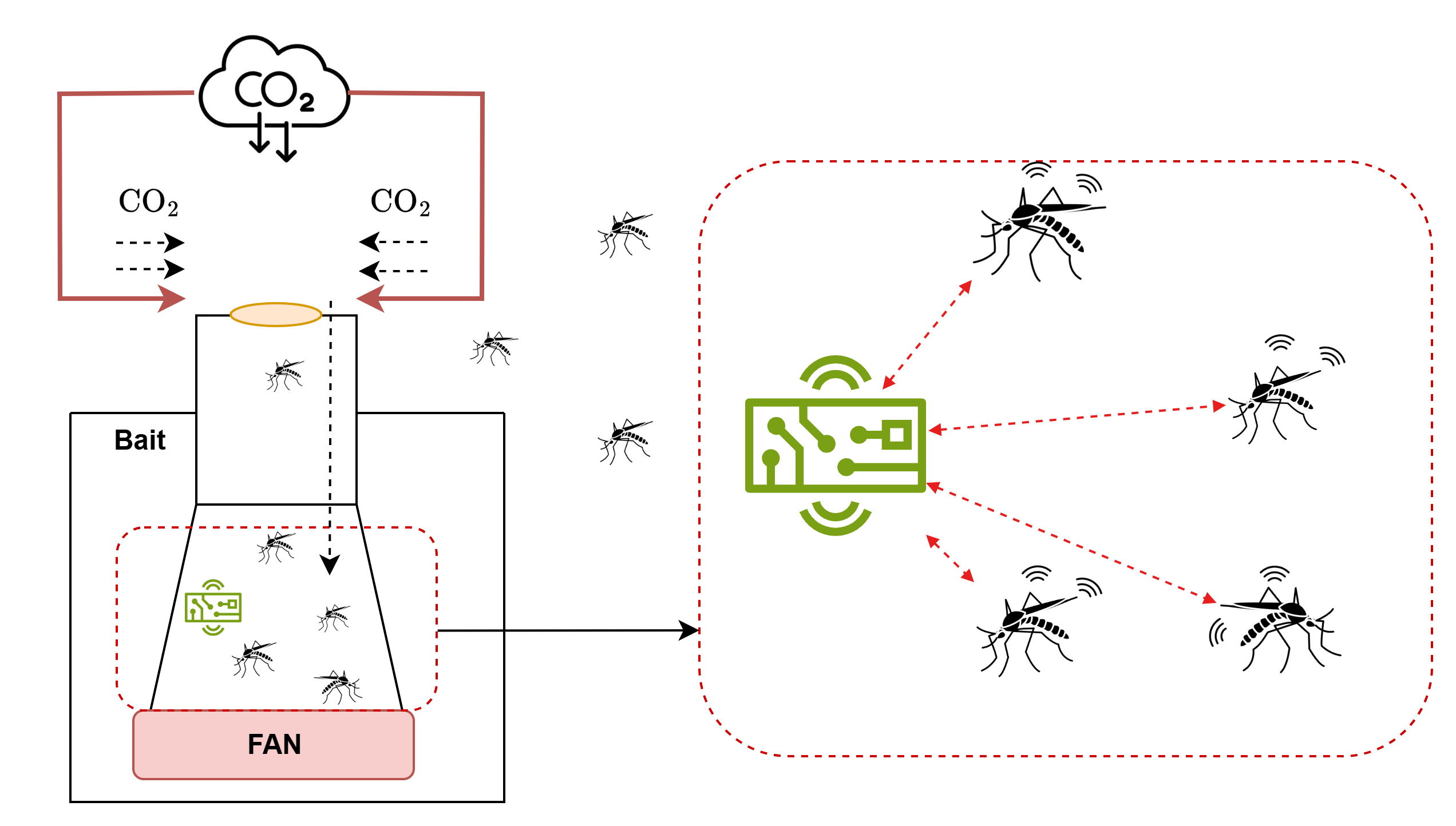}
    \caption{Smart trap system with CO$_2$ bait, audio sensing, and onboard classification.}
    \label{fig:mos-bait}
    \vspace{-0.5cm}
\end{figure}

This work introduces a compact, real-time mosquito detection and classification system tailored for edge deployment. It combines low-power audio sensing with a lightweight dual-output neural network for both presence and species prediction (Figure~\ref{fig:mos-bait}).

\section{Related Work}
\label{sec:related}

\subsection{Audio-Based Mosquito Monitoring}

Traditional mosquito monitoring systems focused on extracting wingbeat frequencies under clean conditions, often limited to single-species detection. Kiskin et al.~\cite{kiskin2019} pioneered the use of convolutional neural networks (CNNs) on spectrograms, demonstrating high accuracy in controlled settings.

The HumBug Project~\cite{humbug2022} introduced large-scale, real-world mosquito audio data collected via smartphones, enabling broader applicability. Fernandes et al.~\cite{fernandes2020} achieved state-of-the-art binary classification using mel spectrograms and compact residual CNNs. More recently, Ramos et al.~\cite{ramos2023acoustic} developed a low-power audio recording system capable of real-time mosquito detection on embedded hardware.

\subsection{Feature Representations}

Early systems used fast Fourier transform (FFT) and mel-frequency cepstral coefficients (MFCCs), but these were sensitive to noise. Mel spectrograms proved more robust and better suited for CNNs. Wavelet-based methods such as the continuous wavelet transform (CWT) improved time-frequency localization. Yang et al.~\cite{yang2019} showed that CWT enhanced detection in noisy environments. However, many approaches still rely on hand-crafted features rather than end-to-end learning.

\subsection{Deep Learning Architectures}

CNNs remain the most common choice for mosquito audio classification. Kiskin et al.~\cite{kiskin2019} and Fernandes et al.~\cite{fernandes2020} used CNNs to classify mosquito species with high precision. Wang et al.~\cite{wang2020crnn} introduced a convolutional recurrent neural network (CRNN) for modeling temporal dynamics.

To support edge deployment, lightweight models have been proposed. Toledo et al.~\cite{toledo2021lstm} achieved 96\% accuracy with an LSTM model containing fewer than 70k parameters. Fernandes et al.~\cite{fernandes2020} also optimized residual CNNs for TinyML platforms.

\subsection{Noise Robustness}

Environmental noise remains a major challenge. Supratak et al.~\cite{supratak2024} addressed this with MosquitoSong+, incorporating noise augmentation and attention mechanisms to maintain high performance under real-world interference. Domain adaptation techniques have also been proposed, though they often require extensive annotated data.

\subsection{Mosquito Audio Datasets}

Several datasets support mosquito audio research:

\begin{itemize}
    \item \textbf{HumBugDB}~\cite{humbug2022}: Over 20 species recorded in real-world conditions with background noise and variable quality.
    \item \textbf{Abuzz}~\cite{abuzz}: Community-sourced smartphone recordings; large-scale but noisily labeled.
    \item \textbf{Curated sets}~\cite{kiskin2019, fernandes2020}: Focused on 2--3 species in clean lab settings.
\end{itemize}

These datasets often lack recordings of overlapping species or swarms. Synthetic datasets have emerged to simulate multi-species scenarios and support mosquito counting tasks under controlled conditions.

\subsection{Limitations in Existing Work}

Despite progress, current systems face several limitations: Limited support for multi-species detection and mosquito counting; few models optimized for embedded devices; scarcity of datasets with realistic noise and swarming conditions.
\subsection{Contribution of This Work}

This study addresses the above gaps by proposing a dual-headed CNN model capable of a scalable training with a synthetic dataset simulating
swarms; multi-species detection; efficient execution on resource-constrained embedded devices; robustness to various noise conditions via augmentation.

Our design builds upon prior work~\cite{kiskin2019, fernandes2020, supratak2024}, while introducing novel contributions in architecture design, data synthesis, and deployment strategy.

\section{Problem Statement}
There is an urgent need for surveillance tools that can estimate mosquito density and distinguish vector species in real time under field conditions. While audio-based detection systems show promise, they are often fragile in noisy environments or when encountering multiple overlapping mosquito sounds.

Additionally, deep learning models used for acoustic insect classification are typically too resource-intensive for low-power embedded devices. Most existing systems are designed for presence detection only, lacking the capability to handle species-level classification from real-world, noisy swarm recordings.

\section{Methodology}

This section outlines the methodology for mosquito sound classification and counting. It includes audio preprocessing, synthetic swarm generation, model architectures, training, and evaluation.

\subsection{Audio Preprocessing and Feature Extraction}

Raw audio signals are sampled at 16 kHz and normalized. To convert 1D waveforms into deep learning-compatible formats, multiple signal transforms are considered:

\subsubsection{Fast Fourier Transform (FFT)}
The FFT is used for global spectral analysis:
\begin{equation}
X[k] = \sum_{n=0}^{N-1} x[n] e^{-j 2\pi kn/N}, \quad k = 0, 1, ..., N-1
\end{equation}
While useful for identifying dominant frequencies, FFT lacks temporal localization.

\subsubsection{Short-Time Fourier Transform (STFT)}
To handle non-stationary wingbeat signals, STFT provides time-frequency representations:
\begin{equation}
X(t, \omega) = \int_{-\infty}^{\infty} x(\tau) w(\tau - t) e^{-j\omega \tau} d\tau
\end{equation}
where $w(\cdot)$ is a window function. STFT forms the basis for spectrogram computation.

In this study, 64-bin log-mel spectrograms computed via STFT are used. Spectrograms are saved as RGB images for compatibility with visual backbones. Noise augmentation improves robustness. To prepare the acoustic features for classification, all mosquito swarm recordings were resampled to a target sampling rate of 16 kHz and then transformed into two alternative time–frequency representations. In the short-time Fourier transform branch, mel spectrograms were computed using a 512-point FFT with a 25 ms Hann window and a 10 ms hop size. Each spectrogram contained 128 mel bands, expressed in decibel scale relative to the signal maximum, and subsequently rendered as an image of 224 × 224 pixels.
\begin{figure}[h]
\centering
\includegraphics[width=0.9\linewidth]{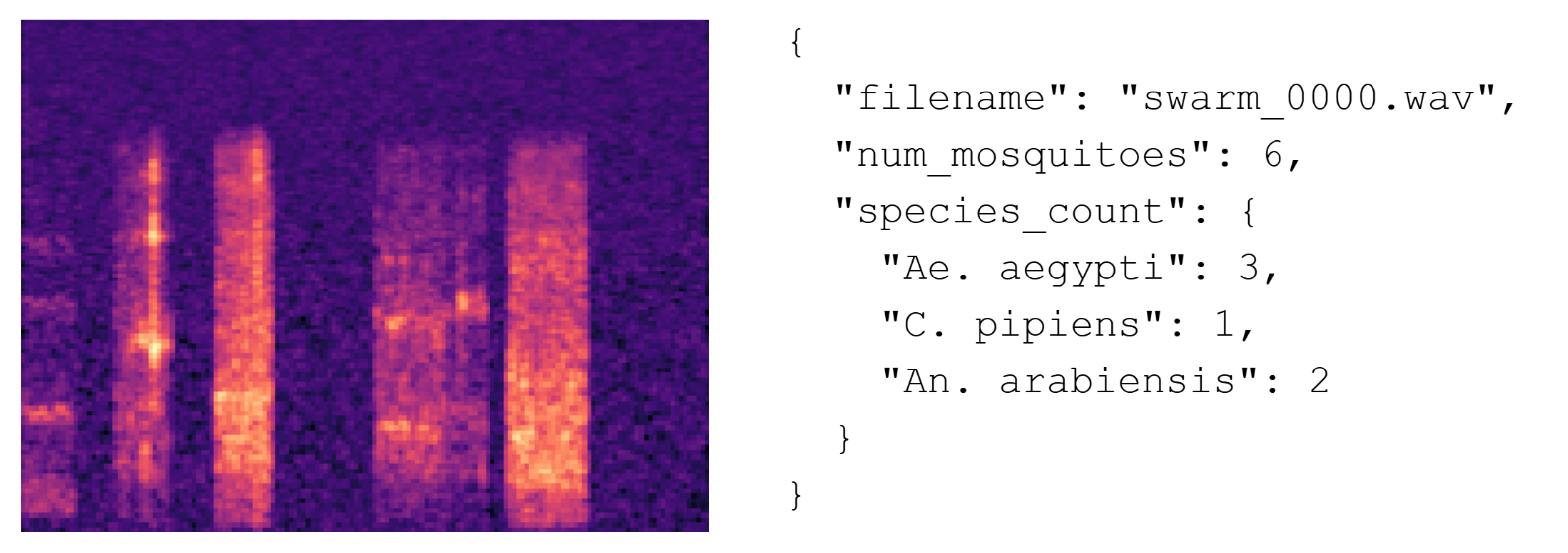}
\caption{Mel-spectrogram of a synthetic mosquito swarm. Vertical lines indicate wingbeat frequencies.}
\end{figure}

\begin{figure}[h]
\centering
\includegraphics[width=0.9\linewidth]{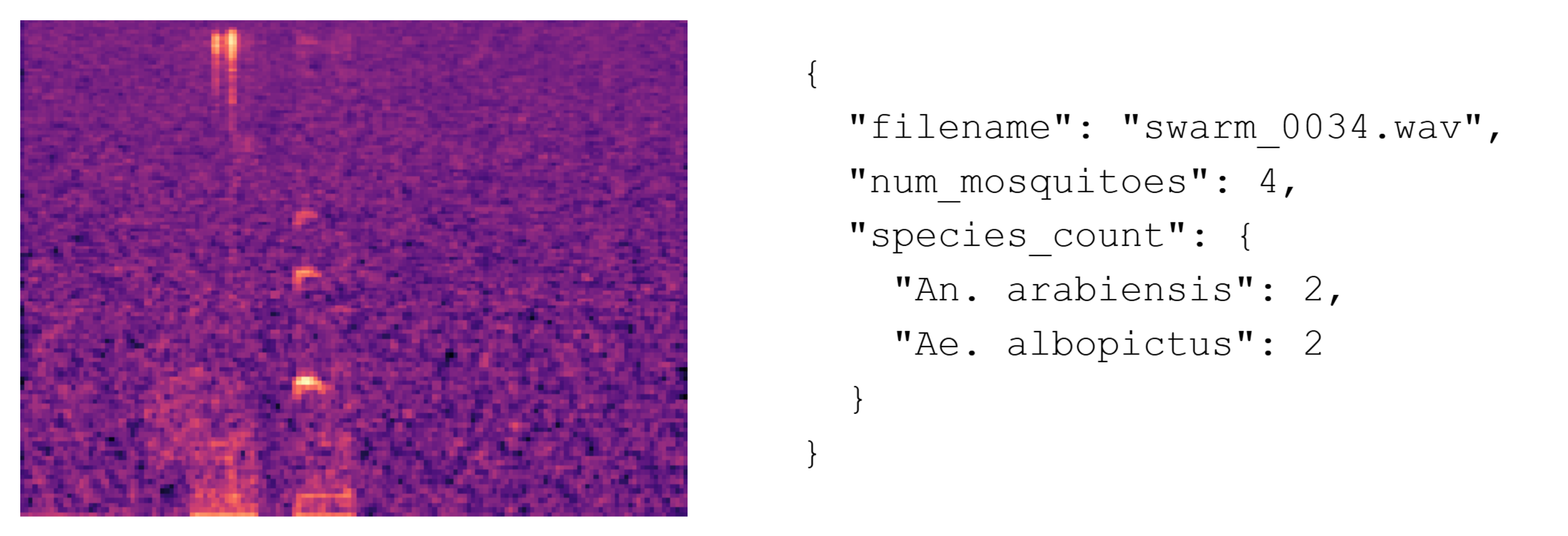}
\caption{Example input with corresponding species label metadata. Used for generating multi-label targets.}
\end{figure}
\subsection{Synthetic Swarm Generation}

Let $\mathcal{S} = {s_1, ..., s_K}$ denote mosquito species. For each synthetic sample, $n \sim \mathcal{U}(1,10)$ mosquitoes are selected:
\begin{equation}
X(t) = \sum_{i=1}^{n} g_i \cdot x_i(t - \tau_i) \cdot \mathbf{1}{[0,T]}(t - \tau_i)
\end{equation}
where $g_i \sim \mathcal{U}(0.2,1.0)$ is gain and $\tau_i \sim \mathcal{U}(0,3.0)$ is time offset. Segments $x_i(t)$ are randomly chunked ($\Delta t \in [0.3, 0.6]$s). White Gaussian noise is added:
\begin{equation}
X{noisy}(t) = X(t) + \epsilon(t), \quad \text{SNR} \in [20, 40] \text{ dB}
\end{equation}
Multi-label ground truth vectors $y \in {0,1}^K$ indicate species presence.

\begin{figure}[H]
    \centering
    \includegraphics[width=0.95\linewidth]{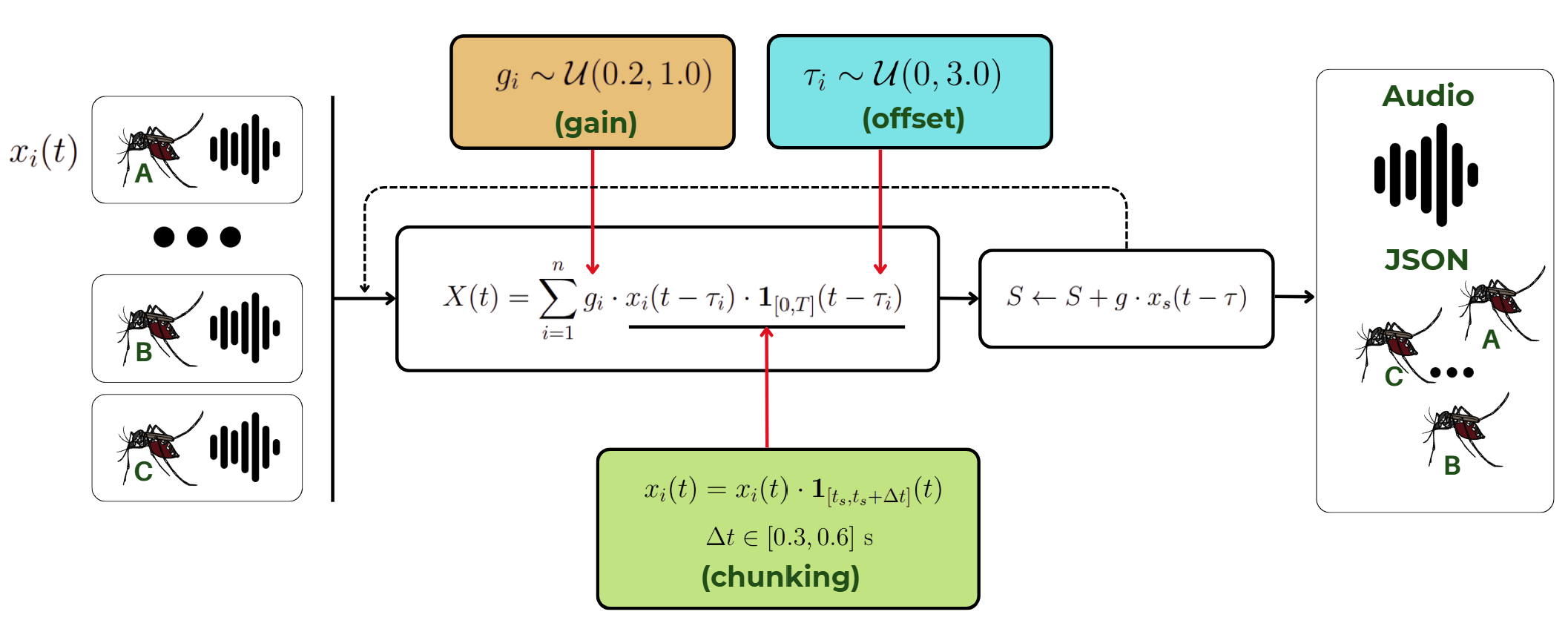}
    \caption{Synthetic swarm generation pipeline.}
    \label{fig:swarm-gen}
\end{figure}

As shown in Figure~\ref{fig:swarm-gen}, swarm synthesis models density, overlap, and timing to create realistic mixtures that support multi-label learning and better generalization. To create the synthetic mosquito swarm dataset, individual mosquito recordings were combined using a carefully controlled strategy that mimicked real-world swarm conditions. Key steps included randomly selecting 3–7 mosquito audio chunks (each 0.3–0.6s) from up to 3 species per sample, applying gain variation (0.2–1.0) to simulate distance, and introducing random time offsets (0–3s) to spread audio events across a 5-second window. This approach ensured biological plausibility and reduced signal artifacts like clipping or redundancy. Additionally, white Gaussian noise was added at varying SNR levels (20–40 dB) to simulate realistic trap environments. Each sample was annotated with species counts and total mosquito numbers for multi-label training. These refinements significantly improved the realism, diversity, and balance of the dataset, enabling more effective and generalizable model training.

\subsection{Dataset}

We use the Abuzz dataset \cite{abuzz}, which contains labeled mosquito wingbeat recordings across 20 species. For this study, we categorize them into three groups:
\begin{itemize}
\item \textbf{Aedes}
\item \textbf{Anopheles}
\item \textbf{Culex}
\end{itemize}
In this study, we focus on Aedes aegypti, Aedes albopictus, Anopheles arabiensis, Anopheles gambiae, Culex quinquefasciatus, Culex pipiens.
\subsection{Model Architectures}

Three neural network architectures are evaluated for multi-label classification:

\subsubsection{CNN (ResNet-18)}
Baseline architecture using ResNet-18 to process $224 \times 224$ RGB spectrograms:
\begin{equation}
\hat{y} = \sigma(W^\top f + b), \quad f = \text{GAP}(\text{CNN}(X))
\end{equation}
Here, $f \in \mathbb{R}^{B \times C}$ is the global average pooled feature representation, and $B$ is the batch size.  
Pros: efficient and edge-friendly. Cons: no temporal modeling.

\begin{figure}[H]
    \centering
    \includegraphics[width=1\linewidth]{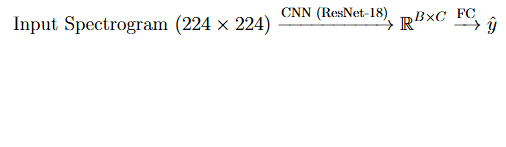}
    \vspace{-2cm}
    \caption{CNN baseline architecture.}
    \label{fig:cnn-arch}
\end{figure}

\subsubsection{CNN + RNN}
CNN features $F \in \mathbb{R}^{B \times C \times H \times W}$ are reshaped and passed to an RNN:
\begin{align}
S &= \text{reshape}(F) \in \mathbb{R}^{B \times T \times C}, \quad T = H \cdot W \\
R &= \text{RNN}(S) \in \mathbb{R}^{B \times d} \\
\hat{y} &= \sigma(W^\top R + b)
\end{align}
Here, $S$ is the sequence embedding of spectrogram patches, and $R$ is the hidden state at the final step.  
This hybrid improves temporal awareness, but can suffer from vanishing gradients.
\begin{figure}[H]
    \centering
    \includegraphics[width=1\linewidth]{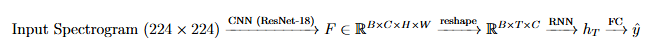}
    \caption{CNN + RNN architecture.}
    \label{fig:cnn-rnn-arch}
\end{figure}
\subsubsection{CNN + LSTM}
Replacing the RNN with an LSTM improves long-range dependency modeling:
\begin{align}
(h_t, c_t) &= \text{LSTM}(S) \\
\hat{y} &= \sigma(W^\top h_T + b)
\end{align}
\begin{figure}[H]
    \centering
    \includegraphics[width=1.1\linewidth]{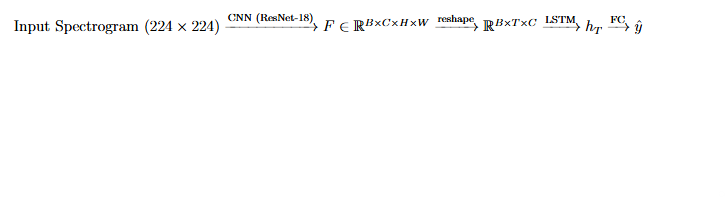}
    \vspace{-2cm}
    \caption{CNN + LSTM architecture.}
    \label{fig:cnn-lstm-arch}
\end{figure}
Here, $h_T$ is the final hidden state capturing temporal dependencies across the sequence.  
This hybrid achieves stronger modeling of long-range wingbeat patterns compared to vanilla RNNs.
\subsection{Training Pipeline}
The training procedure includes spectrogram conversion, label encoding, stratified data splitting, and end-to-end model optimization:
\begin{itemize}
\item BCE loss with sigmoid activation
\item Adam optimizer ($\alpha = 10^{-4}$)
\item Early stopping based on validation loss
\end{itemize}
Predictions $\hat{y} \in [0,1]^K$ are thresholded at different $\tau$ for evaluation.
\begin{figure}[H]
    \centering
    \includegraphics[width=0.95\linewidth]{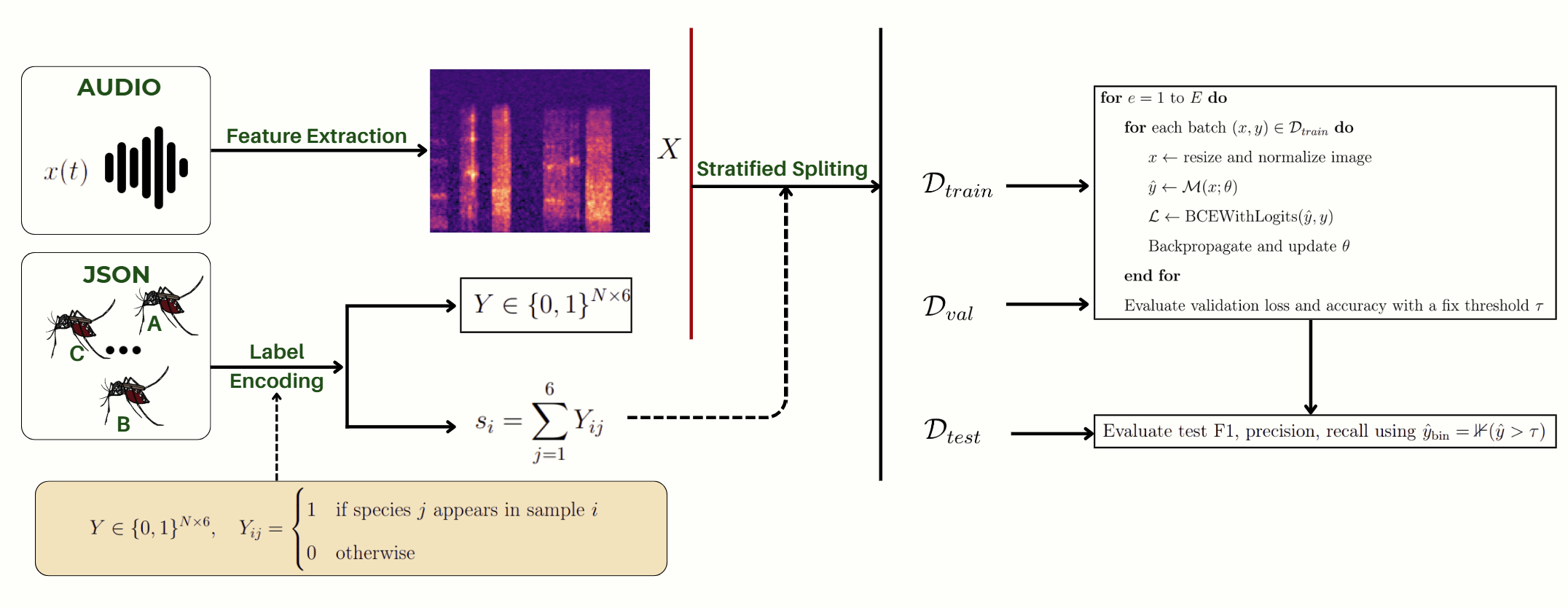}
    \caption{End-to-end learning pipeline for mosquito swarm classification.}
    \label{fig:train-pipeline}
\end{figure}

\begin{algorithm}[H]
\caption{An overview of the end-to-end training pipeline for multi-label mosquito species classification from swarm audio.}

\begin{algorithmic}[1]
\Procedure{TrainSwarmClassifier}{$\mathcal{D}$, Model $\mathcal{M}$, Epochs $E$}    
    \State Convert swarm audio $x(t)$ to STFT spectrogram $X$
    \State Compute binary label matrix $Y \in \{0,1\}^{N \times 6}$ from species counts
    \State Compute label cardinality $s_i \gets \sum_j Y_{ij}$
    \State $\mathcal{D}_{train}, \mathcal{D}_{val}, \mathcal{D}_{test} \gets$ stratified split on $s_i$
    \State Load backbone CNN $\phi$ (e.g., ResNet-18), attach classifier head, or RNN, or LSTM
    \State Initialize optimizer Adam$(\theta)$ and loss function BCE
    \For{$e = 1$ to $E$}
        \For{each batch $(x, y) \in \mathcal{D}_{train}$}
            \State $x \gets$ resize and normalize image
            \State $\hat{y} \gets \mathcal{M}(x; \theta)$
            \State $\mathcal{L} \gets \text{BCEWithLogits}(\hat{y}, y)$
            \State Backpropagate and update $\theta$
        \EndFor
        \State Evaluate validation loss and accuracy with a fix threshold $\tau$
    \EndFor
    \State Evaluate test F1, precision, recall using $\hat{y}_{\text{bin}} = \mathbb{1}(\hat{y} > \tau)$
\EndProcedure
\end{algorithmic}
\end{algorithm}

\subsection{Evaluation Metrics}

We use the following multi-label metrics:
\begin{itemize}
\item \textbf{Multi-label Accuracy}: The average proportion of correctly predicted labels per sample, across all classes. For prediction $\hat{y} \in \{0,1\}^C$ and true label $y \in \{0,1\}^C$, accuracy per sample is: $\frac{1}{C} \sum_j \mathbb{1}(y_j = \hat{y}_j)$
\item \textbf{Macro Precision, Recall, F1-score}
\end{itemize}
These metrics highlight per-class performance and robustness across imbalanced data.

\section{Experimental Results}
\label{sec:result}

This section presents the results of experiments conducted on the multi-label mosquito species classification task using CNN+LSTM, CNN+RNN, and CNN models trained on swarm audio spectrograms. We evaluate performance under different decision thresholds (0.3, 0.5, and 0.7), focusing on the trade-off between sensitivity (recall) and precision, particularly for embedded real-time applications.

\subsection{Impact of Threshold on Model Performance}

\begin{figure}[htbp]
    \centering
    \includegraphics[width=\linewidth]{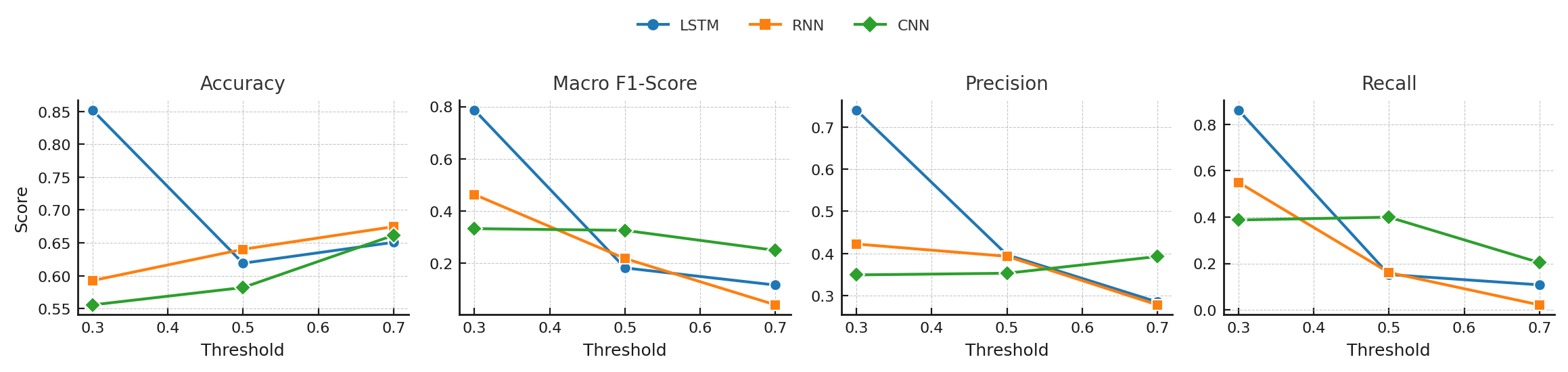}
    \caption{Performance comparison of CNN+LSTM, CNN+RNN, and CNN models across thresholds 0.3, 0.5, and 0.7. Metrics include Accuracy, Macro F1-Score, Precision, and Recall. CNN+LSTM performs best overall, especially in recall and F1 at lower thresholds.}
    \label{fig:per-threshod-performance}
    \vspace{-0.5cm}
\end{figure}

Figure~\ref{fig:per-threshod-performance} compares the performance of CNN+LSTM, CNN+RNN, and CNN across three thresholds using four metrics: Accuracy, Macro F1, Precision, and Recall. The CNN+LSTM consistently outperforms the others, particularly in F1 and recall, confirming its ability to capture long-term audio patterns.

At threshold 0.3, the CNN+LSTM achieves its highest F1 and recall, indicating strong sensitivity to mosquito species presence. However, this comes at the cost of reduced precision due to more liberal predictions. As the threshold increases to 0.5 and 0.7, all models show a decrease in recall and F1 with minimal gains in precision—highlighting the classic trade-off in multilabel classification between detection sensitivity and false positives.

\subsection{Learning Curve Analysis}

\begin{figure}[htbp]
    \centering
    \begin{subfigure}{0.45\linewidth}
        \centering
        \includegraphics[width=\linewidth]{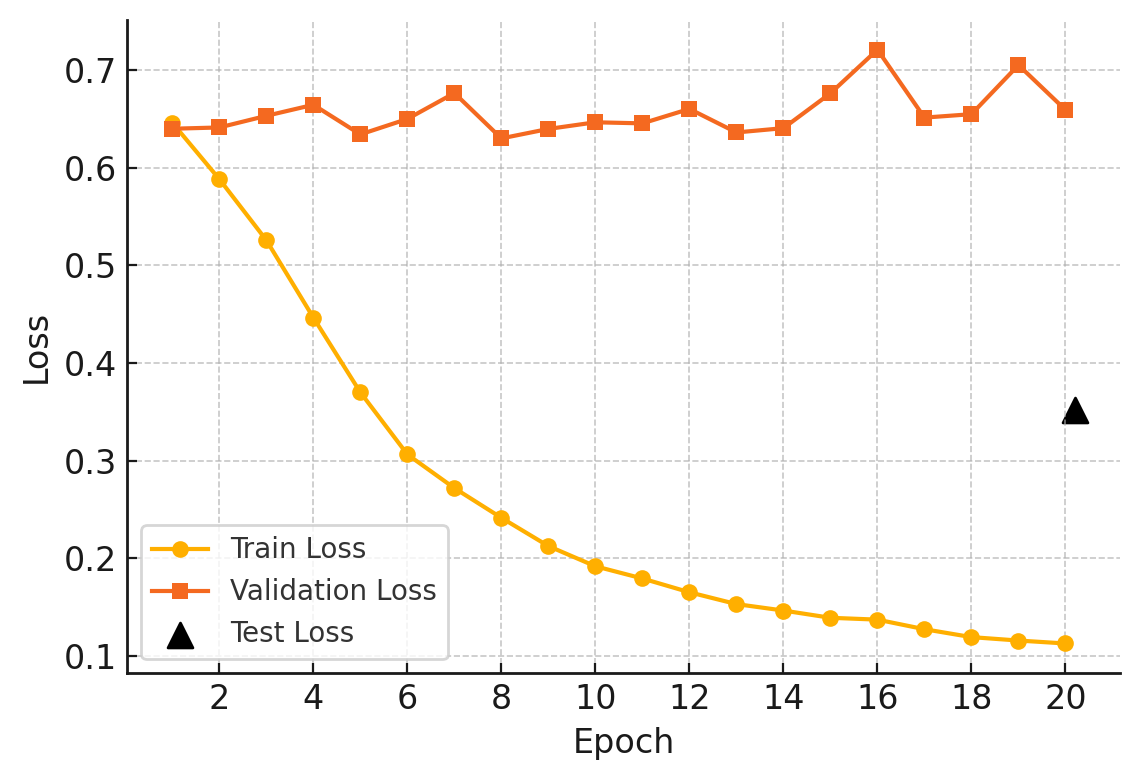}
        \caption{Loss curves at threshold 0.3}
    \end{subfigure}
    \hfill
    \begin{subfigure}{0.45\linewidth}
        \centering
        \includegraphics[width=\linewidth]{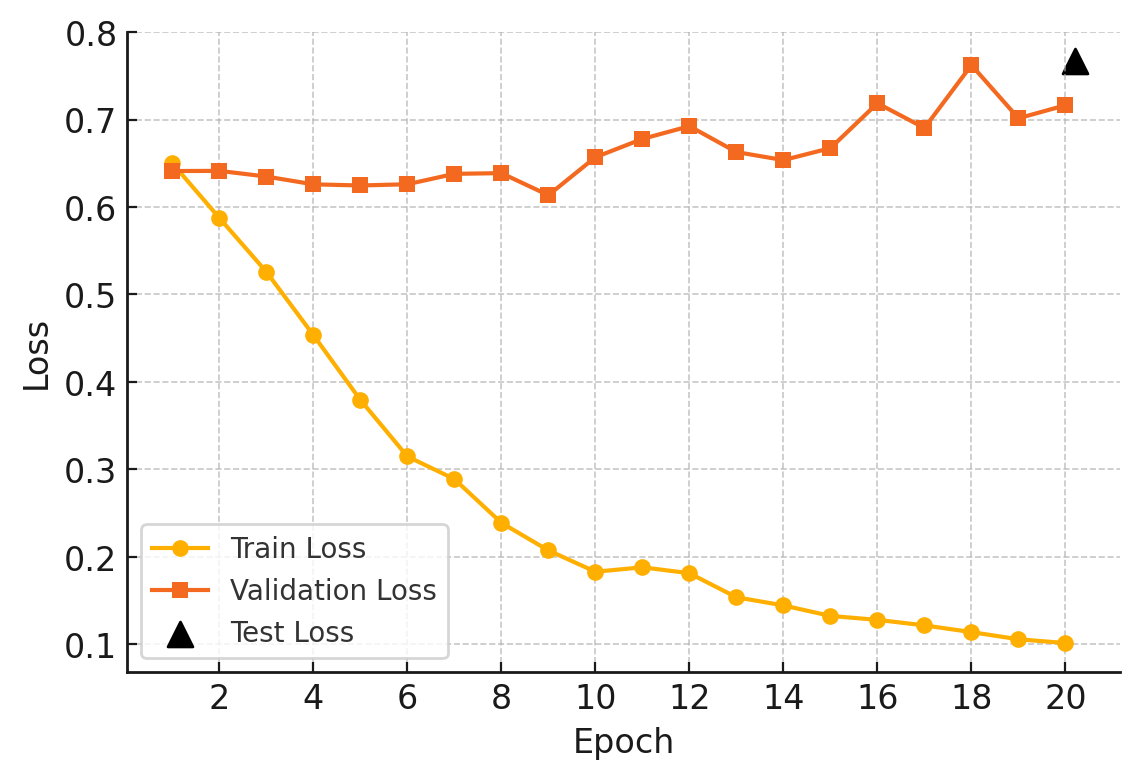}
        \caption{Loss curves at threshold 0.5}
    \end{subfigure}
    \caption{Training and validation loss of CNN+LSTM model under different thresholds. Triangle markers denote final test loss.}
    \label{fig:lstm_threshold_comparison}
\end{figure}

\subsubsection*{Threshold 0.3 – Stable Generalization}

Figure~\ref{fig:lstm_threshold_comparison}(a) shows smooth and consistent convergence in both training and validation losses, suggesting good generalization. The model avoids overfitting, and its final test performance (triangle marker) confirms strong robustness.

\begin{figure}[htbp]
    \centering
    \begin{subfigure}{0.45\linewidth}
        \centering
        \includegraphics[width=\linewidth]{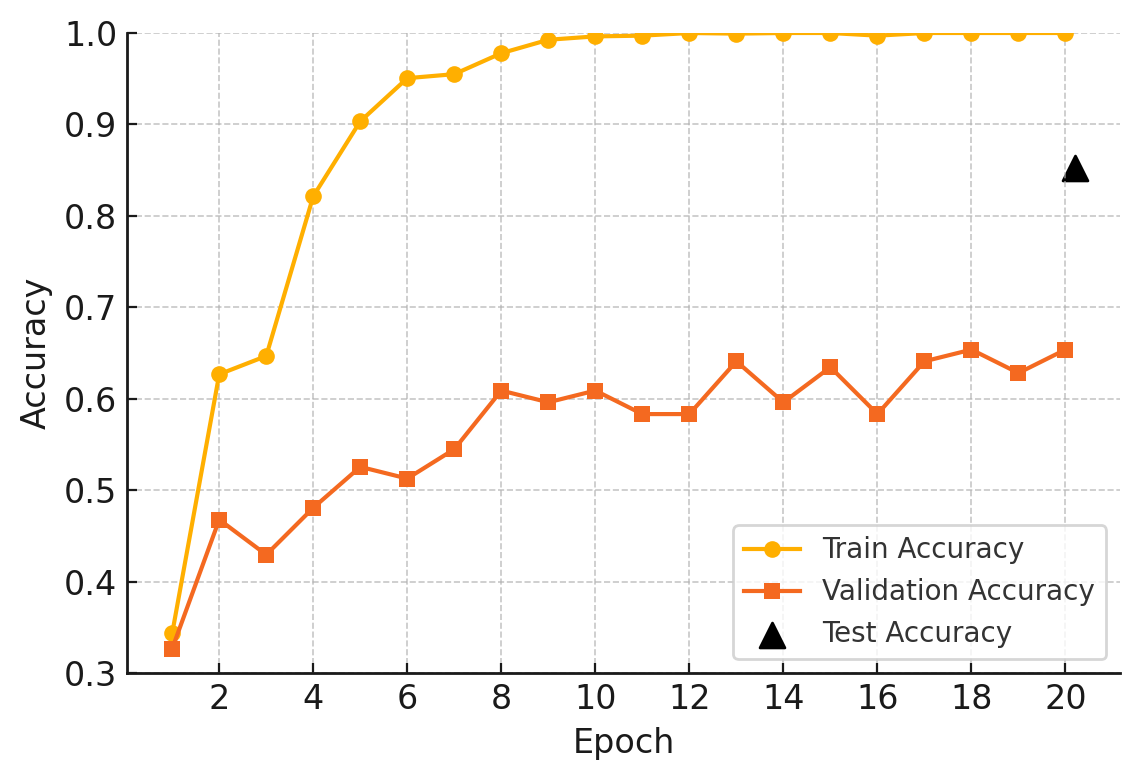}
        \caption{Accuracy at threshold 0.3}
    \end{subfigure}
    \hfill
    \begin{subfigure}{0.45\linewidth}
        \centering
        \includegraphics[width=\linewidth]{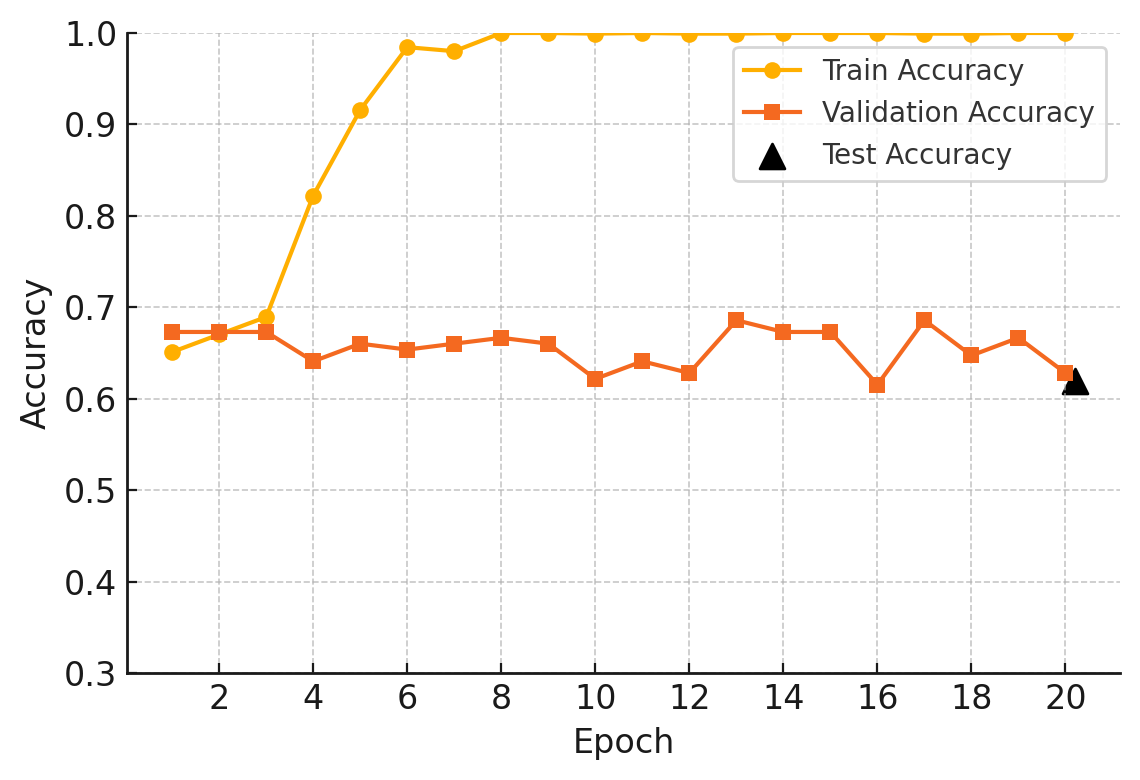}
        \caption{Accuracy at threshold 0.5}
    \end{subfigure}
    \caption{Training and validation accuracy of CNN+LSTM under different thresholds. Triangle markers indicate test accuracy.}
    \label{fig:lstm_acc_threshold_comparison}
\end{figure}

In Figure~\ref{fig:lstm_acc_threshold_comparison}(a), validation accuracy continues to rise after training accuracy saturates—evidence of well-balanced learning. This supports using threshold 0.3 for applications where missing a mosquito is more costly than a false positive, such as in early-warning traps.

\subsubsection*{Threshold 0.5 – Overconfidence and Reduced Sensitivity}

In contrast, Figure~\ref{fig:lstm_threshold_comparison}(b) shows early convergence in training but unstable validation loss, indicating reduced generalization. Figure~\ref{fig:lstm_acc_threshold_comparison}(b) further shows plateaued validation accuracy. The final test performance drops compared to the 0.3 case.

This suggests that threshold 0.5 leads to a model that is overconfident yet less sensitive to minority classes—an undesired outcome for imbalanced multilabel detection tasks.




\subsection{Field Validation in Vung Tau, Vietnam}

\begin{table}[htbp]
\centering
\caption{Summary of mosquito species detections in Vung Tau, Vietnam.}
\begin{tabular}{l c}
\toprule
\textbf{Species} & \textbf{\# Positive Detections (out of 12)} \\
\midrule
\textit{Aedes aegypti} & 10 \\
\textit{Anopheles arabiensis} & 8 \\
\textit{Aedes albopictus} & 6 \\
\textit{Culex quinquefasciatus} & 1 \\
\textit{Anopheles gambiae} & 0 \\
\textit{Culex pipiens} & 1 \\
\bottomrule
\end{tabular}
\label{tab:vungtau-results}
\end{table}

To evaluate real-world performance, we conducted a field validation study in Vung Tau, Vietnam, where swarm audio data was collected using a baited mosquito trap equipped with CO\textsubscript{2} as an attractant to simulate human presence. The trap was designed to encourage natural mosquito swarming behavior while minimizing environmental noise. A high-sensitivity microphone was positioned near the trap to capture continuous flight-tone recordings of approaching and hovering mosquitoes. These real swarm audio samples were then processed using our trained CNN+LSTM detection model, demonstrating the best performance in prior evaluations. During inference, a conservative detection threshold of 0.3 was applied to the model's output probabilities to balance sensitivity and specificity, particularly under field conditions. Based on this setup, the following results summarize the model's performance across six target mosquito species.

Table~\ref{tab:vungtau-results} summarizes detection outcomes from field recordings. There are 12 swarm audios in total.
The model produced strong detections for \textit{Ae. aegypti} (10/12) and \textit{An. arabiensis} (8/12). 
\textit{Ae. albopictus} was detected in half of the samples (6/12) with moderate confidence (~35\%). 
By contrast, \textit{C. quinquefasciatus} and \textit{C. pipiens} rarely exceeded the detection threshold (1/12 each), 
while \textit{An. gambiae} was never detected (0/12).

Considering the context, the dominant detections of \textit{Ae. aegypti} and moderate signals of 
\textit{Ae. albopictus} align with their known distribution in southern Vietnam, including Vung Tau. 
By contrast, the frequent positive detections of \textit{An. arabiensis} are false, as this African vector does not occur in Vietnam. 
Meanwhile, \textit{An. gambiae} was consistently absent, which is geographically correct, although the discrepancy with its sister species suggests wingbeat misclassification. 
\textit{C. quinquefasciatus} is common in Vietnam, and its scant recognition indicates model under-sensitivity. 
Finally, \textit{C. pipiens} holds little public health significance in the region, making its low detection rate acceptable.

\section{Conclusion}

This paper proposed an end-to-end audio-based system for real-time mosquito species detection and classification, designed for deployment in resource-constrained smart traps. The pipeline integrates synthetic swarm audio generation, spectrogram-based feature extraction, deep learning models, and thresholded multi-label classification to achieve robust species-level prediction in noisy, multi-mosquito environments.

In contrast to simple augmentation methods such as time-shifting and pitch-scaling, swarm synthesis functions at a higher level, simulating overlapping organism interactions as well as changing swarm density. This supports multi-label learning, which helps the model adapt better to actual swarm behaviours. Moreover, the synthetic swarm dataset is a key innovation, as it offers a scalable and reproducible resource for surveillance research, thus minimizing the reliance on expensive field data collection.

We explored multiple architectures, including a lightweight CNN, a CNN+RNN hybrid, and a CNN+LSTM model. Among these, the CNN+LSTM consistently achieved the best classification performance, particularly in terms of macro F1-score and recall, highlighting its strength in capturing temporal dynamics.

Thresholding played a key role in balancing recall and precision. At $\tau = 0.3$, high recall was achieved, suitable for surveillance scenarios where missing a species is costlier than a false alarm. Higher thresholds (e.g., $\tau = 0.7$) yielded better precision but degraded recall, reinforcing the need for application-specific tuning.

We also introduced a stratified label cardinality-based data split to improve training balance. The system’s performance across six mosquito species showed both strengths and trade-offs.

Future research directions include: Integrating attention mechanisms to improve spectral focus. Adopting advanced imbalance-aware augmentation, such as mixup variants tailored for rare classes. \cite{cheng2023smote}; applying pruning and quantization for CNN+LSTM deployment on ultra-low-power hardware; enhancing the swarm generator to simulate complex acoustic conditions; exploring weakly supervised and complementary-label learning to handle ambiguous swarm recordings, approaches such as Intra-Cluster Mixup \cite{mai2025-icm} and VLM-generated complementary-label data \cite{mai2025unexplored} may help leverage partially labeled field data and reduce annotation costs; implementing adaptive, context-aware thresholding strategies.

This work demonstrates that lightweight, real-time mosquito classification via audio is both feasible and practical, offering a scalable foundation for vector surveillance in low-resource environments.

\section*{Acknowledgment}
I would like to express my sincere gratitude to Dr. Nguyen Tuan Cuong and Dr. Vo Bich Hien for their invaluable guidance, insightful feedback, and unwavering support throughout the course of this thesis. Their expertise has been instrumental in shaping the research direction, addressing technical challenges, and refining the overall quality of this work. I am deeply appreciative of their mentorship, encouragement, and dedication to my academic development.

\bibliographystyle{IEEEtran}
\bibliography{references}

@article{kiskin2019,
  title={Bioacoustic detection with deep learning},
  author={Kiskin, Ilya and Kindratenko, Volodymyr and Cox, Stephen J},
  journal={Interspeech},
  year={2019}
}

@misc{humbug2022,
  author={MindFoundry},
  title={The HumBug Project},
  year={2022},
  note={\url{https://www.mindfoundry.ai/blog/humbug-2022}}
}

@article{fernandes2020,
  title={Mosquito detection using CNN on smartphone audio},
  author={Fernandes, Hugo and Batista, Pedro},
  journal={arXiv preprint arXiv:2008.09024},
  year={2020}
}

@inproceedings{cheng2023smote,
  title={From SMOTE to Mixup for Deep Imbalanced Classification},
  author={Cheng, Wei-Chao and Mai, Tan-Ha and Lin, Hsuan-Tien},
  booktitle={International Conference on Technologies and Applications of Artificial Intelligence},
  pages={75--96},
  year={2023},
  organization={Springer}
}

@article{yang2019,
  title={Insect wingbeat classification using CWT},
  author={Yang, Min and Zhang, Lei},
  journal={Sensors},
  volume={19},
  number={5},
  pages={1123},
  year={2019},
  publisher={MDPI}
}

@article{wang2020crnn,
  title={CRNN for mosquito species identification},
  author={Wang, Y. and Li, T. and Chen, X.},
  journal={ICASSP 2020 - 2020 IEEE International Conference on Acoustics, Speech and Signal Processing (ICASSP)},
  year={2020},
  pages={900--904},
  publisher={IEEE}
}

@article{toledo2021lstm,
  title={LSTM-Based Mosquito Genus Classification Using Their Wingbeat Sound},
  author={Toledo, Rafael and da Silva, Rafael and Souza, João},
  journal={Proceedings of the ACM on Interactive, Mobile, Wearable and Ubiquitous Technologies},
  volume={5},
  number={3},
  pages={1--20},
  year={2021},
  publisher={ACM}
}

@article{supratak2024,
  title={MosquitoSong+: Noise-robust mosquito classification in real-world environments},
  author={Supratak, Akara and Rattanatamrong, Piyawat and Chomphan, Suphattharachai},
  journal={PLOS ONE},
  volume={19},
  number={2},
  pages={e0310121},
  year={2024},
  publisher={Public Library of Science}
}

@misc{abuzz,
  author = {Abuzz Project},
  title = {Citizen Science to Track Mosquitoes Using Smartphones},
  note = {\url{https://abuzz.stanford.edu/}},
  year = {2018}
}

@inproceedings{ramos2023acoustic,
  title={Acoustic Sensor Module for Mosquito Detection and Classification},
  author={Ramos, Kim and Guico, Maria Leonora C. and Galicia, Jan Kevin A.},
  booktitle={2023 9th International Conference on Computer and Communication Engineering (ICCCE)},
  pages={126--131},
  year={2023},
  publisher={IEEE},
  doi={10.1109/ICCCE58854.2023.10246103}
}

@article{breedlove2022deadly,
  title={Deadly, Dangerous, and Decorative Creatures},
  author={Breedlove, Byron},
  journal={Emerging Infectious Diseases},
  volume={28},
  number={2},
  pages={495--496},
  year={2022},
  publisher={Centers for Disease Control and Prevention},
  doi={10.3201/eid2802.AC2802}
}

@article{boyer2018overview,
  title = {An overview of mosquito vectors of Zika virus},
  author = {Boyer, Sébastien and Calvez, Elodie and Chouin-Carneiro, Thais and Diallo, Diawo and Failloux, Anna-Bella},
  journal = {Microbes and Infection},
  volume = {20},
  number = {11-12},
  pages = {646--660},
  year = {2018},
  publisher = {Elsevier},
  doi = {10.1016/j.micinf.2018.01.006},
  url = {www.sciencedirect.com/science/article/pii/S128645791830039X}
}

@misc{mai2025-icm,
      title={Intra-Cluster Mixup: An Effective Data Augmentation Technique for Complementary-Label Learning}, 
      author={Tan-Ha Mai and Hsuan-Tien Lin},
      year={2025},
      eprint={2509.17971},
      archivePrefix={arXiv},
      primaryClass={cs.LG},
      url={https://arxiv.org/abs/2509.17971}
}

@inproceedings{mai2025unexplored,
  title={The Unexplored Potential of Vision-Language Models for Generating Large-Scale Complementary-Label Learning Data},
  author={Mai, Tan-Ha and Ye, Nai-Xuan and Kuan, Yu-Wei and Lu, Po-Yi and Lin, Hsuan-Tien},
  booktitle={Pacific-Asia Conference on Knowledge Discovery and Data Mining},
  pages={90--102},
  year={2025},
  organization={Springer}
}

\end{document}